\documentclass{article}

\usepackage[preprint, nonatbib]{neurips_2018}

\usepackage[utf8]{inputenc} 
\usepackage[T1]{fontenc}    
\usepackage{hyperref}       
\usepackage{url}            
\usepackage{booktabs}       
\usepackage{amsfonts}       
\usepackage{nicefrac}       
\usepackage{microtype}      

\usepackage{amsmath}
\usepackage{algorithm, algorithmic}
\usepackage{amsfonts, amssymb}
\usepackage{multirow}

\title{Diverse Online Feature Selection}

\author{%
  Chapman Siu \\
  Faculty of Engineering and Information Technology\\
  University of Technology Sydney\\
  Sydney, Australia \\
  \texttt{chapman.siu@student.uts.edu.au} \\
  \And
  Richard Yi Da Xu \\
  Faculty of Engineering and Information Technology\\
  University of Technology Sydney\\
  Sydney, Australia \\
}

\begin{document}

\maketitle

\begin{abstract}
Online feature selection has been an active research area in recent
years. We propose a novel diverse online feature selection method based
on Determinantal Point Processes (DPP). Our model aims to provide
diverse features which can be composed in either a supervised or
unsupervised framework. The framework aims to promote diversity based on
the kernel produced on a feature level, through at most three stages:
feature sampling, local criteria and global criteria for feature
selection. In the feature sampling, we sample incoming stream of
features using conditional DPP. The local criteria is used to assess and
select streamed features (i.e.~only when they arrive), we use
unsupervised scale invariant methods to remove redundant features and
optionally supervised methods to introduce label information to assess
relevant features. Lastly, the global criteria uses regularization
methods to select a global optimal subset of features. This three stage
procedure continues until there are no more features arriving or some
predefined stopping condition is met. We demonstrate based on
experiments conducted on that this approach yields better compactness,
is comparable and in some instances outperforms other state-of-the-art
online feature selection methods.
\end{abstract}

\section{Introduction}

Every day more and more domains are increasing the breadth and depth of
their data every year. It becomes critical to find ways to create
compact and interpretable representations of our
data\cite{Guyon2003}. In this paper we focus on the problem of
diverse online feature selection, where diversity is defined in terms of
the features themselves, and online enabled means that the feature
streams may arrived in mini-batch format or stream-wise fashion.

In this paper we will consider the online feature selection problem,
where features flow into the model dynamically, this can be in groups or
one by one. As the features arrive, a feature selection process is
performed. This formulation differs from the typical online learning
problem, where the feature space is assumed to remain constant while new
instances are shown to the model and the weights subsequently
updated\cite{agarwal14a}.

Existing techniques generally do not consider diversity and instead rely
on other measures, whether it be through use of a regularizer,
statistical tests or correlation measures for feature selection. To this
end, we propose an online feature selection approach called
\emph{Diverse Online Feature Selection} (DOFS). Our framework is
composed of three stages: feature sampling, local criteria and global
criteria for feature selection. In the feature sampling, we sample
incoming stream of features using conditional DPP. In the local
criteria, this is used to assess and select features only when they
arrive, we use unsupervised scale invariant methods to remove redundant
features and optionally supervised methods to introduce label
information to assess relevant features. Lastly we use global criteria
which uses regularization methods to select a globally optimal subset of
features. This three stage procedure continues until there are no more
features arriving or some predefined stopping condition is met.

This work makes the following contributions.

\begin{itemize}

\item
  We propose using conditional DPP as a means for selecting diverse
  features from stream of features. In order to do so, we provide a new
  and novel truncated DPP sampling algorithm.
\item
  To evaluate a stream of features, we introduce an unsupervised, scale
  invariant criteria to remove redundant features and supervised
  approach to address the shortcomings of using only DPP for sampling
  the feature stream.
\item
  our proposed \emph{Diverse Online Feature Selection} (DOFS) achieves
  the strong classification results whether working in supervised or
  unsupervised framework
\end{itemize}

The paper is organized into the following sections. Section 2 we will
lay the preliminary foundations and review related approaches to the
online feature selection problem. In section 3, we will introduce our
framework for Diverse Online Feature Selection (DOFS). In section 4 we
will provide experimental results to demonstrate the effectiveness of
DOFS. We will conclude this work in section 5.

\section{Preliminaries and Related
Work}

In this section we first give a review of offline feature selection and
the state-of-the-art online feature selection counterparts.
Representative methods reviewed are Grafting, Alpha-investing, Online
Streaming Feature Selection (OSFS), Online Group Feature Selection
(OGFS). Afterwards, we will provide a review of determinantal point
processes and the feature sampling problem.

\subsection{Feature Selection}

Traditionally, feature selection has been performed in an offline
setting. The feature selection problem can be framed as follows. We are
given a matrix \(X = [x_1, \dots, x_n] \in \mathbb{R}^{d \times n}\)
which has \(n\) instances and \(d\)-dimension feature space
\(F= [f_1, f_2, \dots, f_d] \in \mathbb{R}^d\). The goal of feature
selection is to selection a subset of the feature space such that
\(U \in \mathbb{R}^l\) where \(l\) is the number of desired features,
where in most cases \(l < d\)\cite{wang2015online}. Offline feature
selection is a widely studied topic with many different
reviews\cite{Guyon2003}. Rather than provide a comprehensive review,
we will instead focus on several selected techniques and their online
feature selection counterparts. We will cover feature selection from two
perspectives as a filter and wrapper method. From the filter approach,
we will consider batch approaches using statistical significance and
spectral feature selection as well as their online variants being Online
Streaming Feature Selection and Online Group Feature Selection
respectively. We will also consider the wrapper methods in the batch
setting using regularization and information criterion approaches, as
well as the online variants being grafting and alpha-investing
respectively.

For completeness, the third approach for feature selection is the
embedded method which perform feature selection in the process of
training as they are specific to models. Approaches here could include
decision tress such as CART, which have built-in mechanism to perform
feature selection\cite{Guyon2003}. To the best of our knowledge,
there are not any embedded methods present from an online feature
selection perspective.

\subsubsection{Correlation Criteria and
OSFS}

The first approach uses the filter method, which evaluates features by
certain criterion and select features by ranking their evaluation values
or by some chosen threshold.

One common approach is to consider the correlation related
criteria\cite{Guyon2003} such as mutual information, maximum margin,
or independence criterion. Of particular interest is conditional
independence criterion which is constructed through consideration of
relevance and redundancy of features in terms of condition
independence\cite{Koller1996}. In this setting the process of
labelling a feature to be relevant or redundant is performed using
statistical tests based on conditional independence.

Online Streaming Feature Selection (OSFS) uses this framework of
relevance and redundancy to determine whether incoming features are
added. When a feature arrives, OSFS first analysis correlation with the
label and determines whether the feature is relevant\cite{Wu2010}.
Once a feature is successfully chosen, then OSFS performs redundancy
test to determine if both previous and current features are redundant
and can be removed. In this setting the redundancy is a key component of
OSFS approach.

\textit{Spectral Feature Selection and
OGFS}\label{spectral-feature-selection-and-ogfs}

A similar approach which also uses statistical tests and falls under the
filter method for feature selection is the use of the spectral feature
selection. In spectral feature selection a graph is constructed. From
this graph where the \(i\)th vertex corresponds to
\(\mathbf{x_i} \in X\), with an edge between all vertex pairs. In this
graph construct its adjacency matrix \(W\), and degree matrix \(D\). The
adjacency matrix is constructed differently depending on the supervised
or unsupervised context. In the spectral analysis setting the adjancy
matrix can be the similarity metric of choice \cite{Zhao2007},
\cite{Wang2015}. For example in the unsupervised context this is can
be the RBF kernel function \cite{Zhao2007}, \cite{Wang2015}, or
a weighted sum of correlation metric and rank coefficient metric
\cite{Roffo_2015_ICCV}. Once the appropriate metric is chosen then,
a feature ranking function is used for filtering the features. This
function can change depending on context, and can be constructed. The
choice of this function can be used to determine the statistical
significance of each individual feature using trace ratio criterion
approach\cite{Grave2011}.

To extend Spectral feature selection to the online setting, the Online
Group Feature Selection (OGFS) has been proposed which considers
incoming \emph{groups} of features and applying spectral feature
selection on a group-wise level. This is used to determine the relevancy
over the particular group of features which has been shown to extend
into the online setting.

\subsubsection{Regularization and
Grafting}

The wrapper method, which uses the machine learning algorithm of
interest as a black box to score subsets of features.

Regularization is typically labelled as a wrapper method in the feature
selection framework, meaning that it uses a model algorithm to jointly
build a model and select features. This is typically employed through
both minimizing empirical error and a penalty. In the context of
regularization, the goal is to encourage sparsity on the feature subset.
Regularizer penalties are typically framed as \cite{PerkinsA2003}

$$\Omega_p(\mathbf{\theta}) = \lambda \sum_{i=1}^m \mathbf{\alpha_i} \lvert \mathbf{\theta_i} \rvert ^p $$

where a choice of \(p=1\) is typically chosen to promote sparsity,
commonly referred to as the Lasso penalty.

To alter this framework to an online setting, the grafting algorithm is
used. Grafting is performed on any model which can be subjected to Lasso
regularizer. The idea behind grafting is to determine whether the
addition of a new feature would cause the incoming feature or
alternatively, any existing feature to have a non-zero weight. With a
chosen parameter \(\lambda\), the regularizer penalty is then
\(\lambda \lvert w_j \rvert\). Thus gradient descent will accept a new
incoming feature \(w_j\) if:

\[ \left\lvert \frac{\partial \mathcal{\bar{L}}}{\partial w_j} \right\rvert > \lambda \]

where \(\mathcal{\bar{L}}\) is the mean loss. In other words, if the
reduction in \(\mathcal{\bar{L}}\) outweighs the regularizer penalty
\(\lambda \lvert w_j \rvert\), then the new incoming feature \(w_j\)
will be chosen. If this test is not passed, then the feature is
discarded. As Grafting makes no assumption on the underlying model, it
can be used in both linear and non-linear models.

\subsubsection{Information Criterion and
Alpha-investing}

Another approach to feature selection in the wrapping sense is the usage
of penalized likelihoods. In the context of single pass feature
selection techniques, penalized likelihoods are preferred
\cite{Zhou2006}. This set of approaches can be framed as:

\[-2 \log(\text{likelihood}) + F\]

where the parameter \(F\) indicates how a criterion is to penalize model
complexity directly.

The alpha-investing algorithm \cite{Zhou2006}, makes use the
information in order to determine whether a new incoming stream of
features is considered to be relevant or not. It makes use of the
\emph{change} in \(\text{log-likelihood}\) and is equivalent to a
t-statistic, which means a feature is added to the model if its p-value
is greater than some \(\alpha\). Alpha-investing works through
adaptively controlling the threshold for adding features. This works
through increasing the wealth \(\alpha\) when a feature is chosen to
reduce the change of incorrect inclusion of features. Similarly when a
feature is assessed wealth is ``spent'', which reduces the threshold, in
order to avoid adding additional spurious features.

In contrast to the previous work, we will tackle feature selection
through the use of feature sampling through determinantal point
processes.

\subsection{Determinantal Point
Process}

We begin by reviewing determinantal point processes (DPPs) and
conditional DPP.

A point process \(\mathcal{P}\) on a discrete set
\(\mathcal{Y} = \{ 1, 2, \dots, N \}\) is a probability measure over all
\(2^{\mathcal{Y}}\) subsets. \(\mathcal{P}\) is a determinantal point
process (DPP) if \(\boldsymbol{Y}\) range over finite subsets of
\(\mathcal{Y}\), we have for every \(A \subseteq \mathcal{Y}\)

\[P(\mathcal{A} \subseteq \boldsymbol{Y}) = \text{det}(\mathbf{K}_{\mathcal{A}})\]

where \(K \in \mathbb{R}^{M \times M }\) is a positive semidefinite
kernel matrix, where all eigenvalues of \(K\) are less than or equal to
\(1\). An alternative construction of DPP is defined by \(L\)-ensembles
where \(L_{ij}\) is a measurement of similarity between elements \(i\)
and \(j\), then DPP assigns higher probability to subsets that are
diverse. The relationship between \(K\) and \(L\) has been shown to be
\cite{kulesza2011learning}

\[K = (L + I)^{-1} L\]

Where \(I\) is the identity matrix. Then the choice of a specific subset
\(Y\) is shown to be \cite{kulesza2011learning}

\[\mathcal{P}_L (\boldsymbol{Y} = Y) = \frac{\text{det}(L_Y)}{\text{det}(L+I)}\]

\subsubsection{Conditional Determinantal Point
Process}\label{conditional-determinantal-point-process}

In our situtation, often we would like to sample future unchosen/unseen
points with the additional constraints based on the currently chosen
features. Suppose that we have input \(X\) and set \(\mathcal{Y}(X)\) of
iterms dervied from the input. Then conditional DPP is defined to be
\(\mathcal{P}(\boldsymbol{Y} = Y | X)\) which is a conditional
probability that assigns a probability to every possible subset
\(Y \subseteq \mathcal{Y}(X)\). Then the model will take form

\[\mathcal{P}(\boldsymbol{Y} = Y | X) \propto \text{det}(L_Y(X))\]

DPP have demonstrated its use in discovering diverse sample points which
has found use in applications such as computer vision and document
summarisation \cite{kulesza2011learning}\cite{kulesza2011kdpps}.
In this context we will consider sampling feature vectors.

\begin{algorithm}
\caption{Conditional Feature Sampling using DPP}
\begin{algorithmic}[1]
\renewcommand{\algorithmicrequire}{\textbf{Input:}}
\renewcommand{\algorithmicensure}{\textbf{Output:}}
\REQUIRE Best candidate feature set: $X \in \mathbb{R}^{d\times n}$ , new set of features $G$, reconstruction error $\alpha$
\ENSURE  Sample of features from $G$
\\ \STATE Construct similarity matrix $L$ based on $X \cup G$. 
\\ \STATE Sample features from $G$ conditioning on $X$ using conditional DPP
\end{algorithmic}
\end{algorithm}

Assuming that the similarity matrix and eigenvalues decomposition \(L\)
is provided, DPP sampling has been shown to have complexity \(O(k^3)\)
\cite{NIPS2010_3969} though Markov Chain DPP sampling (under certain
conditions) is linear in time with respect to the size of data
\cite{Li2016}.

As the above algorithm is inherently unsupervised (i.e.~makes no
assumption on the response vector). This sampling approach could easily
be suitable for both supervised and unsupervised problems. Furthermore,
we propose two different approaches for removing redundant features;
first approach in an unsupervised, scale-invariant manner and second in
a supervised way, leveraging the label information to improve the
consistency of the features chosen.

\subsection{Local Criterion}

Feature sampling alone is insufficient to provide suitable subset of
features without redundancy. Although DPP seeks to promote diversity
within its features it may not necessarily remove all redundant
features. Depending on choice of kernels, kernels may not necessarily be
scale invariant and almost never consistent with respect to response. In
order to address both of these concerns, we turn turn to other criteria
to promote further compactness and reduce redundancy in the feature
selection framework; irrespective of the type of kernel chosen.

\subsubsection{Unsupervised Criterion}

In order to address the \emph{scale-invariant} aspect, we turn towards
non-parametric pair-wise tests to remove redundant features, such as the
Wilcoxon signed-rank test\cite{wilcoxon45}. In our scenario, any two
pairs of features can be viewed as a pair of measurements.

If \(N\) is the sample size, and the pairwise measurements are
\(x_i, y_i\) for the \(i\)th measurement for feature \(x\) and \(y\)
respectively, then the test statistic is calculated through first
ranking the pairs by smallest to largest absolute difference,
\(\lvert x_i - y_i \rvert\). Each pair is then given a rank, in this
scenario we will define \(R_i\) to be the rank of the \(i\)th ranked
pair. Then the statistic is calculated as

\[W = \sum_{i=1}^N (\text{sign}(x_i - y_i) R_i)\]

where \(W\) converges to approximately normal distribution, with
\(z\)-score is given by
\[z = W/\left(\sqrt{\frac{N(N+1)(2N+1)}{6}}\right)\]

Here we propose Wilcoxon signed-rank test to remove any incoming
features which are redundant compared with the present features.

\begin{algorithm}
\caption{Wilcoxon Criterion}
\begin{algorithmic}[1]
\renewcommand{\algorithmicrequire}{\textbf{Input:}}
\renewcommand{\algorithmicensure}{\textbf{Output:}}
\REQUIRE Best candidate feature set: $X \in \mathbb{R}^{d\times k}$, proposed single new feature $f$, significance level $\alpha$
\ENSURE  Boolean, if feature $f$ is discarded or kept
\\ \FOR {each feature $x$ in $X$}
\STATE $p \leftarrow$ Wilcoxon signed-rank test. 
\STATE If $p > \alpha$. Then discard $f$ and terminate, otherwise continue
\ENDFOR
\STATE Keep feature $f$
\end{algorithmic}
\end{algorithm}

As the Wilcoxon signed-rank test requires sorting along a vector of size
\(d\), and all other computation are simple arithmetic, then a single
test will have complexity \(O(d\log (d))\), as this test is repeated
\(k\) times under the proposed Wilcoxon criterion, then it has
complexity \(O(kd \log (d))\).

Although redundancy would already be minimised due to the nature of DPP
sampling, Wilcoxon signed-rank test will provide an approach to removing
redundant features which will help augment the existing approach through
addressing \emph{scale-invariant} aspect which would have been missed.
In addition to using this criteria to detect and remove redundant
features in a scale invariant way, it is also worthwhile to incorporate
information relating to our label in order to select features that are
consistent with our label.

\subsubsection{Supervised Criterion}

Another approach is to make use of information embedded in our label
vector \(Y\). Under this situation it would help address
\emph{consistency} aspect which DPP alone would fail to account for.

Our criteria is based class separability critera in conjunction with
trace ratio criterion \cite{Nie2008} and criterions devised by Wang
et al. (2015). We will define the selected feature set to be \(U\),
\(S_w\) to be the within class scatter matrix and \(S_b\) to be the
between class scatter matrix. There are several ways for class
separability to be defined:

First it can be defined using the mean and variance measures of the
class\cite{Mitra2002}:

\[
\begin{aligned}
S_w(U) &= \sum_{j=1}^c \pi_j \sigma_j\\
S_b(U) &= \sum_{j=1}^c (\mu_j - M_o)(\mu_j - M_o)^T\\
M_o(U) &= \sum_{j=1}^c \pi_j \mu_j\\
\end{aligned}
\]

Where \(\pi_j\) is the priori probability that a pattern belongs to
class \(y_j\), \(U\) is the current candidate feature vector, \(\mu_j\)
is the sample mean vector of class \(y_j\), \(M_o\) is the sample mean
vector for the enture data point, \(\sigma_j\) is the sample covariance
matrix of class \(y_j\).

Similarly it can be constructed through the use of any kernel to define
measure of similarity\cite{Liu2016}:

\[
\begin{aligned}
S_w(U) &= \frac{1}{c}\sum_{j=1}^c \frac{1}{N_j^2} \left( \sum_{k=1}^{N_j} \sum_{l=1}^{N_j} || x_k^{(j)} - x_l^{(j)} ||^2 \right)\\
S_b(U) &= \frac{2}{c(c-1)}\sum_{i=1}^c \sum_{j=1, j\neq i}^c \frac{1}{N_i N_j} \left( \sum_{k=1}^{N_j} \sum_{l=1}^{N_j} || x_k^{(j)} - x_l^{(j)} ||^2 \right) \\
\end{aligned}
\]

Where \(c\) represents the total number of classes for the supervised
classification problem.

Furthermore class separability can also be defined using the label
information directly\cite{wang2015online}:

\[
\begin{aligned}
S_w(U) &= \begin{cases}
               \frac{1}{n}-\frac{1}{n_c}               & y_i = y_j = l\\
               \frac{1}{n} & \text{otherwise}
           \end{cases}  \\
S_b(U) &= \begin{cases}
               \frac{1}{n_c}               & y_i = y_j = l\\
               0 & \text{otherwise}
           \end{cases} 
\end{aligned}
\]

Where \(n_c\) represents the number of instances in class \(c\).

Using any of the between and within class separation criteria defined
above, we can use use these to determine whether a feature is
informative or not. The feature level criterion we will define based on
a single feature \(f\):

\[s(f) = \frac{S_b(f)}{S_w(f)}\]

We can extend this to yield a score for a subset of features based on a
subset of features \(U\), where the goal would be to maximise the
following criterion:

\[F(U) = \frac{\text{tr}(S_b(U))}{\text{tr}(S_w(U))}\]

Both of these criterion can be used to select a stream of features.

\textbf{Supervised Criterion 1} \emph{Given \(U\) to be the previously
selected subset, \(x_i\) denoting the newly arrived feature. Then
feature \(x_i\) will be selected if}

\[F(U \cup f) - F(U) > \epsilon\]

where \(\epsilon\) is a small positive parameter.

\textbf{Supervised Criterion 2} \emph{Given \(U\) to be the previously
selected subset, \(f\) denoting the newly arrived feature. Then feature
\(f\) will be selected if it is a significant feature with
discriminative power}

The significance of the feature can be evaluated by \(t\)-test

\[t(f, U) = \frac{\hat{\mu} -  s(f)}{\hat{\sigma}/\sqrt{\lvert U \rvert}}\]

Where \(\hat{\mu}, \hat{\sigma}\) are the sample mean and standard
deviation of scores of all features in \(U\). If the \(t\)-value reaches
the chosen significance level (in experiments conducted here, chosen to
be \(0.05\)) then the feature is assumed to be significant.

\begin{algorithm}
\caption{Supervised Criterion}
\begin{algorithmic}[1]
\renewcommand{\algorithmicrequire}{\textbf{Input:}}
\renewcommand{\algorithmicensure}{\textbf{Output:}}
\REQUIRE Incoming set of features: $U \in \mathbb{R}^{d\times k}$, significance level $\alpha$
\ENSURE  A set $G$, representing the set of selectioned features
\\ \textit{Initialize:} $G = \{\}$
\\ \STATE Sort $U$ according to function $s$
\\ \FOR {each feature $f$ in $U$}
\STATE If $F(G \cup f) - F(G) > \epsilon$ then $G = G \cup f$
\STATE If $t(f, G) > \alpha$ then $G = G \cup f$
\ENDFOR
\\ \STATE Return $G$
\end{algorithmic}
\end{algorithm}

As both of these criterion are in linear time \cite{wang2015online},
then the remaining complexity comes from the construction of the class
separability critera. The class separability critera has different time
complexity depending on the choice of criterion. In the class separation
criterion from Mitra et all (2002), it relies on the construction of a
covariance matrix, with all other operations being simple arithmetic
operations. As the complexity of covariance matrix calculation is
\(O(k^2d)\), this suggests that the criteria is of complexity
\(O(ck^2d)\), as the covariance is needed to be computed for each class,
and dominates this criterion. Similarly for the class separation which
uses the kernel, the time complexity is \(O(c^2 N^2)\). However if we
use class separation criterion which uses the label information
directly, then it would be in linear time as
well\cite{wang2015online}.

In our supervised criterion, we will accept features if they pass either
\textbf{supervised criterion 1} or \textbf{supervised criterion 2}. It
can also be used in conjunction with unsupervised criterion to result in
providing additional representative features.

After the various criterions which are selected is run, we can proceed
with global criterion to remove redundant features both assessed from
the streaming process and previously accepted features.

\subsection{Global Criterion}

Similar to approaches used by Grafting \cite{PerkinsA2003}, we also
use regulariser to remove redundant features after the conditional
sampling step is complete. This approach was also used in OGFS algorithm
under ``inter-group selection'' criteria which used the Lasso
regulariser specifically\cite{wang2015online}. In this setting we
will consider elasticnet implementation as an alternative to using lasso
to promote sparsity. The regularizer penalty is framed as

\[\Omega_p(\mathbf{\theta}) = \lambda \sum_{i=1}^m \mathbf{\alpha_i} \lvert \mathbf{\theta_i} \rvert ^p \]

Where Elasticnet penalty is specifically
\(\alpha_1 \Omega_1 + \alpha_2 \Omega_2\) which is elasticnet, typically
chosen where \(\alpha_1, \alpha_2 > 0\) and \(\alpha_1 + \alpha_2 = 1\)

Similar to the approach taken by Lasso methods, elasticnet can be used
to select features by having some tolerate \(\lambda \geq 0\) in
mind\cite{Zou05}. Without loss of generality, assume that the
coefficient of a predictor for a particular feature \(f\), is
\(\beta_f\), then we will remove a feature if:

\[\lvert \beta_f \rvert < \lambda \]

Using this, we can now form our global criterion.

\begin{algorithm}
\caption{Global Criterion}
\begin{algorithmic}[1]
\renewcommand{\algorithmicrequire}{\textbf{Input:}}
\renewcommand{\algorithmicensure}{\textbf{Output:}}
\REQUIRE Incoming set of features: $U \in \mathbb{R}^{d\times n}$, tolerance level $\lambda$
\ENSURE  A set $G$, representing the set of selectioned features
\\ \textit{Initialize:} $G = \{\}$
\\ \STATE Fit a model with elasticnet regularizer
\\ \FOR {each feature $f$ in $U$}
\STATE If $\lvert \beta_f \rvert \geq \lambda$ then $G = G \cup f$
\ENDFOR
\\ \STATE Return $G$
\end{algorithmic}
\end{algorithm}

\section{Framework for Diverse Online Feature
Selection}\label{framework-for-diverse-online-feature-selection}

The framework for online feature selection is as follows. First, assume
the current best candidate subset model matrix
\(G = [x_1, \dots, x_n] \in \mathbb{R}^{d \times n}\), where \(d\) is
the number of selected features and \(n\) is the number of instances.
Let the incoming matrix \(\mathbf{G'}\) be
\(\mathbf{G'} = [x'_1, \dots, x'_n] \in \mathbb{R}^{(d + m) \times k}\),
where \(m\) is the number of newly available features. Without loss of
generality we can assume that
\(\mathbf{G'} \in \mathbb{R}^{(d + m) \times n}\), that is the incoming
feature stream have the same number of instances as the best subset
model matrix. Then the difference between the new batch and best subset
is that the new incoming stream of data contains additional features.

Then the online feature selection problem at each iteration selects the
best subset of features of size \(m'\), where \(0 \leq m' \leq m\).

\[\begin{aligned}
{
\begin{array}{@{}c@{}}{
  \begin{bmatrix}
    \multirow{2}{*}{G} \\ \vphantom{\vdots} 
  \end{bmatrix}}_{d\times n}\\
  \\ \vphantom{\vdots} 
  \end{array}
}&
{
\begin{array}{@{}c@{}}{
  \begin{matrix}
    \vphantom{\vdots} \\ \longrightarrow  \\ \vphantom{\vdots} 
  \end{matrix}}\\
  \\ \vphantom{\vdots} 
  \end{array}
}& \hspace{0.5cm}
{
\begin{array}{@{}c@{}}{
  \begin{bmatrix}
    \multirow{3}{*}{$G^{\prime}$} \\ \vphantom{\vdots} \\ \vphantom{\vdots}
  \end{bmatrix}}_{(d+m)\times n}\\
  \\
  \end{array}
}
&
{
\begin{array}{@{}c@{}}{
  \begin{matrix}
    \vphantom{\vdots} \\ \longrightarrow  \\ \vphantom{\vdots} 
  \end{matrix}}\\
  \\ \vphantom{\vdots} 
  \end{array}
}& \hspace{0.5cm}
{
\begin{array}{@{}c@{}}{
  \begin{bmatrix}
    \multirow{2}{*}{$G^{\prime\prime}$} \\ \vphantom{\vdots} 
  \end{bmatrix}}_{(d+m')\times n}\\
  \\ \vphantom{\vdots} 
  \end{array}
}
\end{aligned}\]

If the initial best subset was size \(d\) and there were an additional
\(m\) features available to be selected, the online feature selection
algorithm will then select \(d+m'\) features.

\subsection{Diverse Online Feature
Selection}\label{diverse-online-feature-selection}

\begin{algorithm}
\caption{Diverse Online Feature Selection}
\begin{algorithmic}[1]
\renewcommand{\algorithmicrequire}{\textbf{Input:}}
\renewcommand{\algorithmicensure}{\textbf{Output:}}
\REQUIRE Best feature candidate feature set $X$, Feature stream $F$, label vector $Y$
\ENSURE  A set $G$, representing the set of selectioned features
\\ \WHILE{features are arriving}
\STATE $G \leftarrow$ generate new group of features
\\ \STATE Sample features from $G$ using DPP to get sampled subset $G'$ conditional on $X$
\\ \FOR {$f$ in $G'$}
\STATE \textbf{Local Criterion}: Evaluate feature $f$ using unsupervised and/or supervised criterions to determine relevancy
\\ \STATE \textbf{Global Criterion}: Perform redundacy check based on regulariser
\ENDFOR
\ENDWHILE
\end{algorithmic}
\end{algorithm}

As the complexity of the various parts have been touched on in the
previous sections, we can put them all together to get the overall
complexity of DOFS. If a single iteration has the best candidate feature
set to be \(G \in \mathbb{R}^{d \times n}\), with a stream of new data
of size \(F \in \mathbb{R}^{(d+m)\times n}\). Then the complexity of DPP
sampling will be, \(O((d+m)^3)\) where \(m\) represents the number of
features available to be selected from the feature stream after DPP
sampling. The unsupervised criterion will then have complexity at most
\(O((d+m)\log (d+m))\) and supervised criterion will have complexity at
most \(O(cn^2 (d+m))\) or as little as being linear in time.

Overall the worse case complexity will be
\(O((d+m)^3) + O(cn^2 (d+m)) = O(\max((d+m)^3, cn^2 (d+m))\). Where
\(n\) represents the number of incoming instances used to update our
feature selection, and \(m\) is the number of new available features. If
we use the class separation criteria which has linear time complexity,
then the overall complexity will reduce to DPP sampling, i.e.
\(O((d+m)^3)\).

\section{Experiments}\label{experiments}

Various experiements were conducted to validate the efficiency of our
proposed method. We used several benchmark datasets, several other
state-of-the-art online feature selection methods are used for
comparison including Grafting, OSFS, and OGFS. The classification
accuracy, log-loss and compactness (the number of selected features) are
used to measure performances of the algorithms in our experiments.

We divide this section into three sub-sections, including introduction
to our data sets, the experimental setting and the experimental
comparisons.

\subsection{Benchmark Data Sets}

The benchmark datasets are from UCI Machine Learning Repository, and the
Micro Array datasets. The information of these datasets are described in
the table below.

\begin{table}
\centering
\begin{tabular}{|l|l|l|}
\hline
Data Set & \#instances & \#dim. \\ \hline
Ionosphere & 351 & 34 \\ \hline
Spambase & 4601 & 57 \\ \hline
Spectf & 267 & 44 \\ \hline
Wdbc & 567 & 30 \\ \hline
Colon & 62 & 2000 \\ \hline
Leukemia & 72 & 7129 \\ \hline
Lung Cancer & 181 & 12533 \\ \hline
Prostate & 102 & 12600 \\ \hline
\end{tabular}
\end{table}

There are four datasets from UCI repository (Ionosphere, Spambase,
Spectf, Wdbc), and four datasets from microarray dataset (colon,
leukemia, lung cancer, prostate).

\subsection{Experimental Settings}\label{experimental-settings}

In our experiments, Grafting and OGFS used elasticnet setup with
\(\lambda = 0.15\) for the regularizer penalty and intergroup selection
parameters respectively. For OSFS, OGFS, DOFS the threshold parameter
\(\alpha\) is set to \(0.05\).

To simulate online group feature selection, a similar setup by Wang et
al. was followed. The group structures of the feature space was
simulated by dividng the feature space as a global feature stream by
streaming features in groups of size \(m\). In our experiements we set
\(m \in [5, 10]\) as suggested by Wang et al.. Models were compared
using using existing Matlab implementations such as the LOFS
library\cite{Yu2016}, whilst the DOFS implementation was completed
in Python using scikit-learn library. The DOFS models include the
unsupervised variant (without consideration of class separability), and
supervised variant using the criteria which used the label information
directly.

\subsection{Experimental Results}


\begin{table}[h]
\centering
\begin{tabular}{|l|l|l|l|l|}
\hline
\multirow{2}{*}{Data Set} & \multicolumn{2}{l|}{Alpha-investing} & \multicolumn{2}{l|}{OSFS} \\ \cline{2-5} 
 & \#dim & accu. & \#dim & accu. \\ \hline
Ionosphere & 10 & 87.18 & 8 & 79.93 \\ \hline
Spambase & 45 & 77.18 & 54 & 60.99 \\ \hline
Spectf & 7 & 79.40 & 5 & 79.09 \\ \hline
Wdbc & 21 & 71.53 & 10 & 62.74 \\ \hline
Colon & 4 & 79.76 & 4 & 85.48 \\ \hline
Leukemia & 16 & 66.67 & 5 & 91.83 \\ \hline
Lung cancer & 69 & 86.67 & 7 & 83.43 \\ \hline
Prostate & 25 & 97.09 & 5 & 91.84 \\ \hline
\end{tabular}
\end{table}

\begin{table}[h]
\centering
\begin{tabular}{|l|l|l|l|l|}
\hline
\multirow{2}{*}{Data Set} & \multicolumn{2}{l|}{Grafting} & \multicolumn{2}{l|}{OGFS} \\ \cline{2-5} 
 & \#dim & accu. & \#dim & accu. \\ \hline
Ionosphere & 32 & 91.76 & 26 & 88.26 \\ \hline
Spambase & 50 & 92.28 & 24 & 91.07 \\ \hline
Spectf & 37 & 80.36 & 5 & 71.27 \\ \hline
Wdbc & 24 & 94.82 & 18 & 96.07 \\ \hline
Colon & 26 & 84.26 & 102 & 90.47 \\ \hline
Leukemia & 13 & 94.53 & 63 & 100 \\ \hline
Lung cancer & 19 & 96.53 & 33 & 99.44 \\ \hline
Prostate & 17 & 95.53 & 96 & 98.00 \\ \hline
\end{tabular}
\end{table}

\begin{table}
\centering
\begin{tabular}{|l|l|l|l|l|l|l|}
\hline
\multirow{2}{*}{Data Set} & \multicolumn{2}{l|}{\begin{tabular}[c]{@{}l@{}}DOFS\\(DPP only)\end{tabular}} & \multicolumn{2}{l|}{\begin{tabular}[c]{@{}l@{}}DOFS\\(Unsupervised)\end{tabular}} & \multicolumn{2}{l|}{\begin{tabular}[c]{@{}l@{}}DOFS\\(Supervised)\end{tabular}} \\ \cline{2-7} 
 & \#dim & acc. & \#dim & accu. & \#dim & accu. \\ \hline
Ionosphere & 11 & 87.75 & 9 & 88.12 & 23 & 86.47 \\ \hline
Spambase & 24 & 86.44 & 10 & 82.54 & 37 & 88.26 \\ \hline
Spectf & 10 & 79.40 & 31 & 79.26 & 29 & 79.57 \\ \hline
Wdbc & 8 & 86.29 & 13 & 86.62 & 13 & 86.34 \\ \hline
Colon & 750 & 90.32 & 47 & 95.70 & 38 & 94.52 \\ \hline
Leukemia & 836 & 63.37 & 5 & 68.41 & 58 & 100 \\ \hline
Lung cancer & 1366 & 94.48 & 7 & 91.08 & 88 & 98.37 \\ \hline
Prostate & 1441 & 78.43 & 42 & 92.02 & 34 & 86.61 \\ \hline
\end{tabular}
\end{table}

\textit{Comparison of DOFS
variants}\label{comparison-of-dofs-variants}

If we consider the three variants of DOFS, the usefulness of both the
supervised and unsupervised algorithms are clearly warranted as if we
consider accuracy to be metric of interest, supervised/unsupervised
variant has better accuracy in 4 of 8 models. However, in the situations
which supervised variant underperforms, the difference with the
unsupervised variant is much lower. In the results above, it is clear
that the unsupervised variant promotes greater compactness over the
supervised variant. This can be thought of as the algorithm allowing
more ``chances'' for a feature to be accepted and passed through the
model. This is further highlighted by the difference when there is no
redundancy check placed as in the variant which only uses DPP. In this
setting there is a distinct possibility that extrenous set of features
is selected despite the use of conditional DPP, which comes at a cost of
performance, as can be observed in all the Micro Array datasets, where
the number of features selected is at least 10 times, and in some cases
100 times larger than the other two variants provided.

Overall from the results above comparing against either the supervised
or unsupervised DOFS algorithm, we can see that DOFS generally has
superior performance compared with Alpha-investing and OSFS algorithms,
whilst it seems to be competitive with Grafting and OGFS. In generally
there is a trade-off between compactness and performance; where it would
perform better than alpha-investing and OSFS algorithm whilst being less
compact, and competitive with Grafting and OGFS whilst having better
compactness. What is interesting is that DOFS algorithm demonstrates
inferior performance against all methods when using the Prostate
dataset.

\textit{DOFS vs Alpha-investing}\label{dofs-vs-alpha-investing}

Both variants of DOFS manages to outperform alpha-investing in 6 of the
8 datasets. Excluding Prostate dataset, in the ionosphere the
performance is within 2\%. When comparing compactness, alpha-investing
is generally more compact. Overall DOFS (Unsupervised) has roughly
\textasciitilde{}5-7\% improvement and DOFS (Supervised)
\textasciitilde{}8-10\% improvement over alpha-investing approach for
online feature selection. In terms of compactness, the unsupervised
variant has even better compactness for 5 of the 8 datasets chosen,
demonstrating that the unsupervised variant of DOFS consistently
outperforms alpha-investing both in terms of accuracy and compactness.
Overall our algorithm is able to select sufficient features with
discriminative power.

\textit{DOFS vs OSFS}\label{dofs-vs-osfs}

Unsupervised and supervised variant of DOFS outperforms OSFS in 7 of the
8 datasets, with roughly \textasciitilde{}4-6\% improvement for the
unsupervised variant and \textasciitilde{}10\% for the supervised
variant. OSFS achieves greater compactness in all combination of
datasets and variants of DOFS algorithm, with the exception of
unsupervised DOFS and Spambase dataset. This demonstrates the trade-off
in compactness of representation in this algorithm against the accuracy
in performance. Overall our algorithm is able to select sufficient
features with discriminative power.

\textit{DOFS vs Grafting}\label{dofs-vs-grafting}

Across the board Grafting appears to be a superior algorithm in terms of
accuracy. Unsupervised DOFS outperforms Grafting in only 1 of the 8
datasets, whilst supervised variant outperformed Grafting in 3 of the 8
datasets. On average the difference in accuracy for the supervised
variant suggests that we suffer \textasciitilde{}1-2\% loss in accuracy,
demonstrating minimal loss in performance. With this in mind, in 4 of
the 5 datasets where performance was worse than grafting, the supervised
DOFS achieved improved compactness by \textasciitilde{}30\%.

\textit{DOFS vs OGFS}\label{dofs-vs-ogfs}

Compared with OGFS, DOFS unsupervised variant outperforms in 2 of 8
datasets and DOFS supervised outperforms in 3 of 8. On average the
difference in accuracy for the supervised variant suggests that we
suffer \textasciitilde{}1-2\% loss in accuracy on average, demonstrating
minimal loss in performance. Given this trade-off, supervised variant of
DOFS manages to have an improved compactness by \textasciitilde{}12\%.
This demonstrates that DOFS is a competitive algorithm retaining similar
level of performance whilst promoting further compactness.

\section{Conclusion}

In this paper, we have presented a new algorithm called DOFS which can
select diverse features both in a supervised or unsupervised
environment. We have explored the limitations of using DPP for feature
sampling alone, and demonstrated the necessity and value of introducing
additional redundancy checks to provide a competitive performance. This
framework allows us to efficient select features that arrive by groups
and also one by one. We have divided online feature selection into three
stages: DPP sampling, local criteria and global criteria. We have
designed several criteria for selecting the optimal number of \(k\) to
sample from DPP, trace rank approach for supervised learning problem,
group wilcoxon signed rank test and Lasso to reduce redundancy.
Experiments have demonstrated that DPP is on par or better than other
state-of-the-art online feature selection methods whilst being more
compact through the use of the UCI and Micro Array benchmark datasets.

\section{Acknowledgment}\label{Acknowledgment}

We would like to acknowledge everyone in the data science team at Suncorp Group Limited for their help and support in making this possible.
Any opinions, findings, conclusions or recommendations expressed in
this material are those of the author(s) and do not
necessarily reflect the views of Suncorp Group Limited.



\bibliographystyle{plain}
\bibliography{main}

\end{document}